\documentclass[11pt]{article}

\usepackage[preprint]{acl}

\usepackage{times}
\usepackage{latexsym}

\usepackage[T1]{fontenc}

\usepackage[utf8]{inputenc}

\usepackage{microtype}

\usepackage{inconsolata}

\usepackage{graphicx}
\usepackage{booktabs}
\usepackage{amsmath} 
\usepackage{booktabs}
\usepackage{placeins}
\usepackage{dblfloatfix}
\title{Explanations of Large Language Models Explain Language Representations in the Brain}

\author{
 \textbf{Maryam Rahimi\textsuperscript{1}} \quad
 \textbf{Mohammad Reza Daliri\textsuperscript{1,3}} \quad
 \textbf{Yadollah Yaghoobzadeh\textsuperscript{2,4}}
\\
\textsuperscript{1} School of Electrical Engineering, Iran University of Science and Technology\\
\textsuperscript{2} School of Electrical and Computer Engineering, University of Tehran\\
\textsuperscript{3} School of Cognitive Sciences, Institute for Research in Fundamental Sciences\\
\textsubscript{4} Tehran Institute for Advanced Studies, Khatam University\\
\texttt{maryamrahimiha@gmail.com,} \quad \texttt{daliri@iust.ac.ir,} \quad \texttt{y.yaghoobzadeh@ut.ac.ir}
 }

\begin{document}
\maketitle
\begin{abstract}

Large Language Model (LLM) representations are known to align with brain activity during language processing, but it remains unclear what drives this alignment. We test whether explainable AI (XAI) can help answer this: using attribution methods, we quantify the contribution of each input word to an LLM's next-word predictions and use these explanations to predict fMRI data from participants listening to narratives. We find that gradient-based attribution methods robustly align with brain activity, contribute unique variance beyond acoustic and word-rate confounds, and outperform internal representations in early auditory regions. Using conductance, we extend attribution from words to individual layers, asking what each layer's attribution reveals about the model's computation and how this relates to its brain alignment. Early layers show greater word-type sensitivity and align preferentially with auditory regions, whereas the final layer's attribution is dominated by positional information and exhibits broad cortical alignment. Together, these findings demonstrate that attribution-based explanations can be used not only to measure LLM–brain alignment but to characterize what it reflects.

\end{abstract}
\section{Introduction}\label{sec1}

\begin{figure*}[t]
    \centering
    \includegraphics[width=1\linewidth]{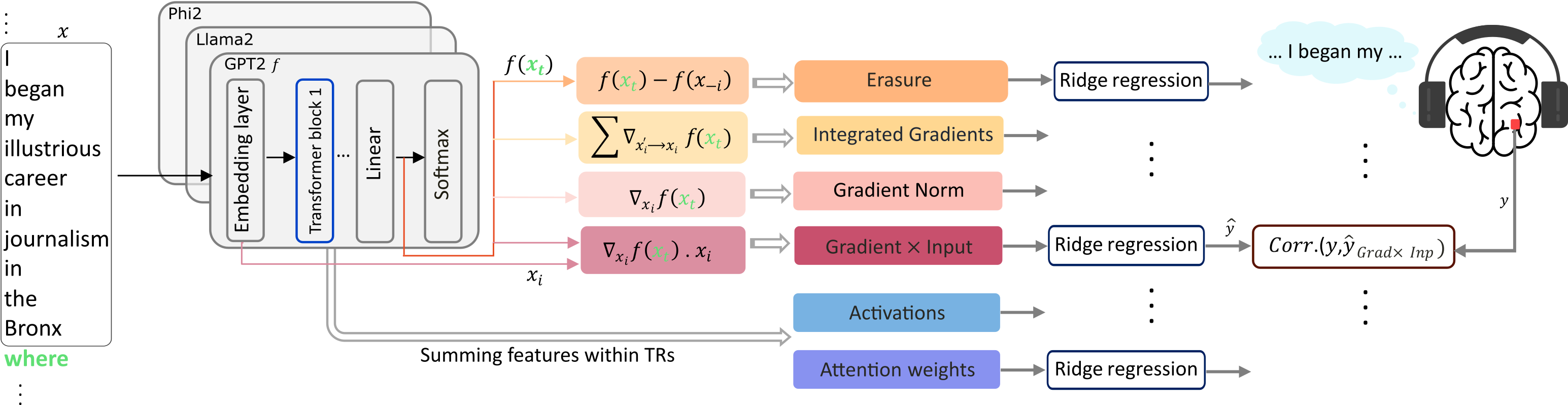}
    \caption{\textbf{The LLM-brain alignment framework.} 
     Explanation scores, extracted from an LLM, are used as features in a ridge regression model to predict fMRI activity. Alignment, or `brain score,' is then quantified as the Pearson correlation between the predicted ($\hat{y}$) and actual brain responses (y).}
    \vspace{-5mm}
    \label{fig:Approach}
\end{figure*}

A growing body of research has shown significant alignment between large language models (LLMs) and human brain activity when both process the same linguistic input \cite{toneva2019, kumar2024shared}. This alignment is typically measured using linear encoding models trained on internal LLM representations (e.g., activations) to predict brain responses to the same stimuli \cite{caucheteux2022brains, schrimpf2021}.

However, we still lack a clear understanding of how these stimulus-related representations form in each system and why they align. Both the brain and LLMs are opaque, black-box systems whose internal representations are high-dimensional, making it unclear whether alignment reflects a shared computational principle or an incidental correlation with some other property of the stimulus \cite{antonello2023, antonello2024predictive, hadidi2026spurious}.

Explainable AI (XAI) offers a set of tools built specifically to open the black box of LLM computation. This motivates our central question: \textbf{do explanations of an LLM's own computation also correspond to activity in the human brain?} We investigate this by focusing on attribution methods, which quantify how much each preceding word contributes to a model's next-word prediction, producing a feature space that is directly interpretable in terms of the model's own computation. If attribution features align with brain activity, their explicit structure offers a way to move beyond the alignment score itself: we can ask \emph{what} the aligned feature space represents about the model, and whether that characterization can help explain \emph{why} it aligns with each brain region.

Addressing these questions would be notable for two communities: for XAI and interpretability researchers, since brain alignment would offer external, biologically-grounded evidence that these methods capture something real about language processing \cite{kar2022interpretability}; and for language neuroscience, since attribution gives a concrete, measurable account of how much weight prior context receives -- a long-standing open question in the field \cite{kumar2024shared}.

To test this, we apply a diverse set of attribution methods across multiple LLM architectures and evaluate their alignment with fMRI activity during naturalistic story listening (Figure~\ref{fig:Approach}). We first ask whether attribution-based explanations align with brain activity beyond what low-level acoustic and word-rate confounds can explain. Second, we compare these attribution-based features with conventional internal representations (activations and attention weights). Finally, using layer conductance, we ask what each layer's attribution reveals about the model's hierarchical computation, and whether this characterization helps explain \emph{where} in the brain that layer aligns.

Our contributions are as follows:
\begin{itemize}
    \item We introduce brain alignment for a range of attribution methods across multiple LLMs.
    \item We show that gradient-based explanations achieve the strongest brain alignment among tested attribution methods and, despite their substantially lower dimensionality, provide unique variance beyond acoustic and word-rate confounds and outperform internal representations in early auditory regions.
    \item We move beyond measuring LLM--brain alignment to characterizing it by using layer conductance, revealing that early and late layers align with the brain through different signals -- lexical content versus positional information.
    \item We find that layers most important to the model's own next-word prediction also tend to align more strongly with brain activity overall, a relationship that holds under most robustness checks.
\end{itemize}

\section{Related Work}
\textbf{LLM–brain alignment}
The alignment between LLMs and brain activity has been widely studied using linear encoding models that map neural responses to model representations. This relationship has been explored across architectures, training regimes, and linguistic performance, with transformer models consistently showing stronger alignment than embeddings or RNNs—often linked to their predictive capabilities \cite{schrimpf2021, caucheteux2022brains, mischler2024contextual}. These findings have been used to probe neural processes such as prediction, semantics, and compositionality \cite{toneva2022combining, antonello2024gemv}, and to evaluate or steer model development \cite{aw2023training, moussa2024improving, AlKhamissi2024TheLL}.
Our work builds on this foundation but shifts focus from internal representations to explanatory dynamics, asking whether the process by which a model attributes importance to input words parallels cortical computation.

\paragraph{XAI for LLM-brain alignment}
A nascent but growing body of work uses XAI to move beyond treating LLMs and the brain as ``black boxes.'' One approach builds new, inherently interpretable encoder models of the brain, for example by replacing opaque LLM features with verbal explanations  or answers to targeted questions \cite{antonello2024gemv,singh2025evaluating}. A concurrent approach analyzes the full encoding pipeline, using attribution to compare which input words are important for the brain alignment task versus the model's own prediction objective \cite{proietti2025finegrainedanalysisbrainllmalignment}. Our work takes a third path, using explanations of the original LLM's computational process as a direct feature space to encode brain activity. A preliminary study by \cite{russo2022} has shown that saliency scores from GPT-2 could predict brain activity. Our study confirms and substantially expands upon this finding by systematically evaluating a broader range of attribution methods across multiple LLMs and brain regions. More importantly, we leverage this framework for mechanistic discovery, investigating how the functional role of each model layer in its predictive process corresponds to its alignment with the brain's processing hierarchy.

\paragraph{LLM-brain alignment for XAI}
A key challenge in explainable AI (XAI) is determining whether explanations meaningfully improve human understanding of model behavior. While algorithmic metrics assess model faithfulness, evaluating cognitive plausibility relies on behavioral human studies. However, behavioral evaluation suffers from human biases—such as confirmation bias—and severe practical constraints \cite{rim2024human, biessmann2021turingtesttransparency, doshivelez2017rigorousscienceinterpretablemachine, alufaisan2021explainable, colin2022human}. Consequently, recent work advocates integrating neuroscientific insights into interpretability research \cite{kar2022interpretability, he2024multilevelinterpretabilityartificialneural, mineault2025neuroaiaisafety}.
Positioned at this intersection, we propose brain alignment not as a standalone metric of explanation quality, but as an external characterization signal and a biological-plausibility probe. Under the hypothesis that neural alignment provides a proxy for cognitive plausibility \cite{kar2022interpretability}, our empirical mapping offers a neurobiologically grounded framework with potential advantages in objectivity and ecological validity for contextualizing interpretability artifacts.

\section{Methods}
\label{sec:methods}
Our pipeline follows a standard model-encoding practice for measuring brain alignment: we expose language models to the same stories presented to the participants during neuroimaging, extract features from the models, and use those features to linearly predict the recorded brain activity (Figure \ref{fig:Approach}). The following sections detail each component of this pipeline.

\subsection{fMRI Datasets and Preprocessing}
\label{sec:datasets}
We used three story-listening datasets from the Narratives fMRI collection \citep{nastase2021narratives}: ``Pieman'', ``Shapes'' (the shapessocial condition), and the consecutive stories ``Slumlord'' and ``Reach for the Stars One Small Step at a Time''. 
We used the preprocessed unsmoothed cortical-surface data provided by the Narratives release, which were processed using the fMRIPrep pipeline and projected onto the fsaverage surface. No additional preprocessing or spatial smoothing was applied in our analyses. Following the quality-control procedure of the original datasets, our analyses included 150 recordings from 147 unique participants. 
All encoding analyses were performed at the voxel level. For regional analyses and visualization, voxels were grouped into anatomically defined regions of interest (ROIs) using the Destrieux atlas \citep{Destrieux2010}.
Additional details regarding participants, preprocessing, recording selection, and ROIs are provided in Appendix~\ref{app:datasets}.

\subsection{Language Models} \label{sec:lms}
We selected three language models spanning a range of scales and architectures. GPT-2 (124M parameters) \cite{Radford2019} was chosen as the primary model because it has an established track record in the brain-alignment literature, from early demonstrations of LLM-brain alignment \cite{schrimpf2021} to recent work continuing to use it as a reference architecture \cite{caucheteux2023, Oota2022JointPO, merlin-etal-2026-brain}. To ensure our findings were not model- or scale-specific, we extended the analysis to two larger, more recent models: Llama 2 (7B parameters) \cite{touvron2023}, an open-source model family competitive with other open-source chat models, and Phi-2 (2.7B parameters) \cite{gunasekar2023}, which achieves performance comparable to significantly larger models on language understanding and commonsense reasoning. All pretrained models were accessed via Hugging Face Transformers \cite{hugging_face}.

\subsection{Feature spaces}
We constructed three classes of feature spaces. Our primary feature spaces are attribution-based explanations. We compare these against two complementary baselines: (i) internal model representations (activations and attention weights) and (ii) an acoustic control baseline used to assess whether attribution explains variance beyond low-level acoustic and temporal properties of the stimulus.

\subsubsection{Attribution-Based Feature Spaces}
\label{sec:attribution}
\paragraph{Word Attribution}

Feature attribution methods are a class of XAI techniques that quantify the contribution of each input token to a model's next-word prediction.
We employ four distinct attribution methods: two gradient-based methods, Gradient Norm \cite{simonyan2013deep_gradn} and Gradient $\times$ Input \cite{denil2015}; a path-based method, Integrated Gradients \cite{sundararajan2017axiomatic}; and a perturbation-based method, Erasure. 

\textit{Gradient Norm:} This method computes the L1 norm of the gradient of the model’s output with respect to the input token embeddings. The L1 norm ensures that the importance scores reflect the magnitude of the gradient, providing a straightforward measure of each token's influence on the model's prediction \cite{simonyan2013deep_gradn, li2015visualizing_grad_n}.

\textit{Gradient $\times$ Input:} This method \cite{denil2015} multiplies the gradient of the model’s output with the input embeddings. To derive per-token scores, we computed the L2 norm of the product, an approach that has been found to perform well in tasks requiring output explainability \cite{atanasova2020diagnostic}.

\textit{Integrated Gradients:} This gradient-based technique \cite{sundararajan2017axiomatic} addresses the issue of saturation by accumulating gradients along a path from a baseline (e.g., a zero vector) to the actual input. For this method, we used the \texttt{Captum} library \cite{captum_software}.

\textit{Erasure:} This method \cite{li2016understanding_erasure} estimates the importance of an input token by evaluating the effect of its removal or masking on the model's prediction, directly quantifying the token's criticality by observing the impact of its absence.

\paragraph{Construction of Attribution Feature Spaces}
To compute LLM attributions, we segmented the input story into overlapping sequences using a sliding window with stride one. While sliding windows are common in LLM-brain alignment work, applying this design before extracting attributions lets us obtain multiple attribution scores per word -- one for each relative position it occupies across windows. Considering these scores together gives a more complete picture of a word's contribution to the output than a single score would, which could otherwise simply reflect how close the word is to the prediction \cite{rahimi2026layerwisepositionalbiasshortcontext}.

The window length was chosen based on prior work showing that embeddings of upcoming words can be reliably decoded from brain activity up to a second before they are perceived \cite{goldstein2022}. We used a sliding window spanning approximately one second of linguistic input, containing 10 context words. This window length has also been used in relevant prior work \cite{toneva2019, Lammarre2022}.

For each input sequence, we computed importance scores for every token, with respect to the actual next word in the story following that window (target-conditioning details in Appendix~\ref{app:target_conditioning}). 
Each attribution method outputs a vector per token; we derived a single per-token score via the $L_1$ norm ($L_2$ for Gradient$\times$Input, following prior work showing this yields more interpretable scores \cite{atanasova2020diagnostic}). Tokens belonging to the same word were summed to a unified word-level score, allowing us to align our feature space with the fMRI sampling rate using word-level timing.

Since windows overlap, each word is processed within up to 10 windows; we stored these as a vector associated with that word. Words appearing in fewer than 10 windows (the first and last 10 words of the story) were excluded to ensure consistent processing, producing a feature matrix of dimensions $(W-20) \times 10$, where $W$ is the total number of words in the story. 
We note that, because a word's aggregated vector combines its contribution across window occurrences up to ten words before its target, this feature space can carry information about words that occur later in the story.
\paragraph{Layer Conductance}\label{sec:conductance}
Conductance is a gradient-based attribution method that extends Integrated Gradients to measure how attribution flows through a network's intermediate units \cite{dhamdhere2018}. A prior study applied Layer Conductance to examine how positional information is distributed across LLM layers \cite{rahimi2026layerwisepositionalbiasshortcontext}; we follow the same conductance procedure here, computing attributions with \texttt{Captum} over the MLP sublayer of each transformer block, targeting the logit assigned to the actual next word in the story (Target conditioning and baseline details Appendix~\ref{app:target_conditioning}).

To construct the layer conductance feature space, we employed a moving-window approach similar to that used for feature attribution. For each window, we compute a vector of conductance scores corresponding to each model layer for every word. Each element in this vector quantifies the contribution of a specific model layer to the model’s next-word prediction given that word as input. Aggregating these scores across all windows in which a word appeared resulted in a per-layer feature representation for the entire story.

\textit{Layer Importance:}
We also use conductance to derive a profile of how the LLM’s layers are used during next-word prediction; these layer-importance scores are not used as features in the brain-alignment framework. For each word, we average its conductance values across the windows in which it appears, within each layer, yielding one conductance value per word. The word’s ``dominant'' layer is the layer with the highest averaged conductance. Aggregating over all words yields, for each layer, the share of words for which it is dominant—this is the layer-importance profile.

\subsubsection{Internal Representation Baselines}
\label{sec:baselines}
We constructed two baseline feature spaces from the models' internal representations: activations and attention weights.

\paragraph{Activations}
We extracted contextual hidden-state representations from a single pre-selected layer for each model, following previous LLM--brain alignment studies \cite{goldstein2022,caucheteux2022brains,schrimpf2021,caucheteux2023}: layer 8 for GPT-2 and layer 12 for Llama~2 and Phi-2, which have previously been reported to yield the strongest brain alignment \cite{caucheteux2022brains,mischler2024contextual}.
Using a sliding window, each word was processed together with its preceding context, where the context length was set to the model's maximum input size (1024 tokens for GPT-2 and Llama~2, and 2048 tokens for Phi-2). From each input sequence, we retained the hidden state corresponding to the final token, yielding one contextual representation for each predicted word. Hidden-state vectors belonging to subword tokens were summed to obtain word-level representations.

\paragraph{Attention Weights} 
For constructing the attention feature space, we followed the procedure of \citet{Lammarre2022}. Using a sliding-window procedure analogous to that used for attribution features, we fed the model input sequences of 11 words (10 preceding words and the target word). We extracted the attention maps from all heads and layers and computed the column-wise mean of each attention matrix to obtain a single attention vector for the input sequence. Attention values corresponding to subword tokens were summed to obtain word-level representations. The resulting vectors were then concatenated across all attention heads and layers, producing one feature vector per word with dimensionality equal to the number of layers × attention heads × input-sequence length.

\subsubsection{Acoustic Control Baseline}
\label{sec:acoustic_baseline}

Recent studies have shown that low-level confounds, particularly acoustic properties and word rate, can account for a substantial portion of the brain predictivity of speech and language models \cite{shimizu2025interpretableembeddingsspeechenhance,hadidi2026spurious}. To evaluate whether attribution features explain variance beyond these confounds, we constructed an acoustic baseline consisting of mel-spectrogram, root mean square (RMS) energy, and word-rate features. Acoustic features were extracted using the \texttt{Librosa} library \cite{McFee2015librosaAA}. Complete details of the acoustic feature extraction are provided in Appendix~\ref{app:acoustic_details}.
\begin{figure*}[htbp]
\centering
\includegraphics[width=\linewidth]{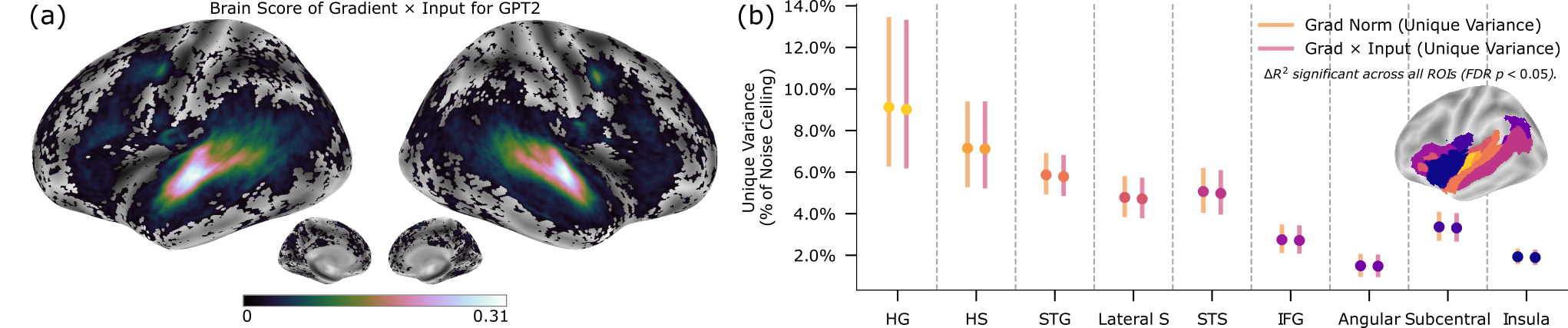}
\caption{
\textbf{Brain alignment and unique variance explained by attribution methods.}
\textbf{(a)} Brain alignment of Grad $\times$ Input explanations. Brain alignment of other models and feature spaces are shown in Figure~\ref{fig:A_brain_maps_all_features}
\textbf{(b)} Unique variance of attribution methods with respect to acoustic and word-rate confounds. Dot colors indicate ROIs as shown in the brain map; error-bar colors denote different feature spaces. (right hemisphere in Figure~\ref{fig:A_unique_var_rh}(b)).}
\vspace{-5mm}
\label{fig:Attribution_predict}
\end{figure*}
\subsection{Brain Encoding}
\label{sec:encoding}
To quantify the alignment between model-derived features and neural activity, we adopted the standard voxel-wise linear encoding framework \citep{caucheteux2022brains,kumar2024shared}. For each participant and voxel, we fitted an independent ridge regression model to predict the corresponding fMRI BOLD response from the model-derived features elicited by the same story. Brain alignment was quantified as the Pearson correlation between predicted and observed BOLD responses on held-out data, referred to throughout the paper as the \emph{brain score} \cite{yamins2014}.  Complete details of feature temporal alignment, the encoding pipeline, hyperparameter selection, and cross-validation are provided in Appendix~\ref{app:encoding_details}.
\paragraph{Acoustic Control via Variance Partitioning}
\label{sec:acoustic_control}
To assess whether attribution features explain variance beyond low-level acoustic information, we used variance partitioning established by \citet{de2017hierarchical}. We fitted two ridge regression models for each voxel using the same encoding pipeline described above: an \emph{acoustic-only} model using the acoustic feature space alone, and a \emph{joint} model that stacks these acoustic predictors with the attribution feature space (independently for Gradient Norm and Gradient $\times$ Input).  We quantify the unique contribution of attributions by subtracting the variance explained ($R^2$) by the acoustic-only model from the $R^2$ of the joint model \cite{shimizu2025interpretableembeddingsspeechenhance}. 
Following standard practice for handling non-predictive fits, we compute these $R^2$ values by clipping any negative fold-averaged Pearson correlations to zero before squaring \cite{tuckute2023many,anderson2024deep,hadidi2026spurious}.


\subsection{Statistics}
\label{sec:statistics}

Brain scores are computed per voxel and per individual, pooling participants across the three narrative datasets. In brain maps, we mark voxels where the average score is significantly different from zero using a Wilcoxon signed-rank test applied independently at each voxel across subjects, with false discovery rate (FDR) correction. In ROI-level figures, we average voxels within each ROI per participant and, where applicable, average across the three language models; we then apply a participant-level bootstrap hypothesis test, either one-sided (significantly above zero) or two-sided (comparing two conditions), also FDR-corrected (Appendix~\ref{app:stats}). All ROI-level scores are reported relative to a noise ceiling estimated from inter-subject correlation (Appendix~\ref{app:methods_noise_ceiling}).

\section{Results}

\subsection{Explanations Align with Brain Activity}\label{sec:attribution_predict}
We first examine the alignment between attribution-based features and brain activity. Attribution significantly predicts activity in a large, bilateral cohort of language-related voxels (Figure~\ref{fig:Attribution_predict}(a); the same pattern was observed across all tested models (Figure~\ref{fig:A_brain_maps_all_features})). Among the tested attribution methods, Gradient Norm and Gradient$\times$Input achieve significantly higher brain alignment than Erasure and Integrated Gradients (Figure~\ref{fig:A_attribution_comparison}); we therefore focus subsequent analyses on these two methods. 

Alignment is highest in Heschl's gyrus (HG) and Heschl's sulcus (HS), reaching approximately 60\% of the noise ceiling across all models (Figure~\ref{fig:A_model_grad_norm_vs_grad_x_input}). Because alignment peaks in primary auditory cortex, we ask whether attribution features capture only low-level temporal or acoustic properties of the stimulus, such as word rate, or carry additional information. Variance partitioning shows that attribution features contribute significant unique variance beyond an acoustic baseline,  not only in auditory cortex but extending into higher-order regions (Figure~\ref{fig:Attribution_predict}(b); right-hemisphere and voxel-wise $\Delta R^2$ maps in Figure~\ref{fig:A_unique_var_rh}). 

To better understand what this additional information reflects, we characterize the attribution feature space directly (Appendix~\ref{app:attribution_conductance_characterization}). The attribution patterns reveal that LLMs rely more strongly on content words than function words and on words occurring at the beginning or end of the context window compared with those in the middle. This is consistent with the unique variance above reflecting more than a purely acoustic signal, though we do not test this relationship directly.

\subsection{Explanations Outperform Internal Representations in Auditory Regions}\label{explanaitions_outperform}
\begin{figure}[!t]
\centering
\includegraphics[width=\linewidth]{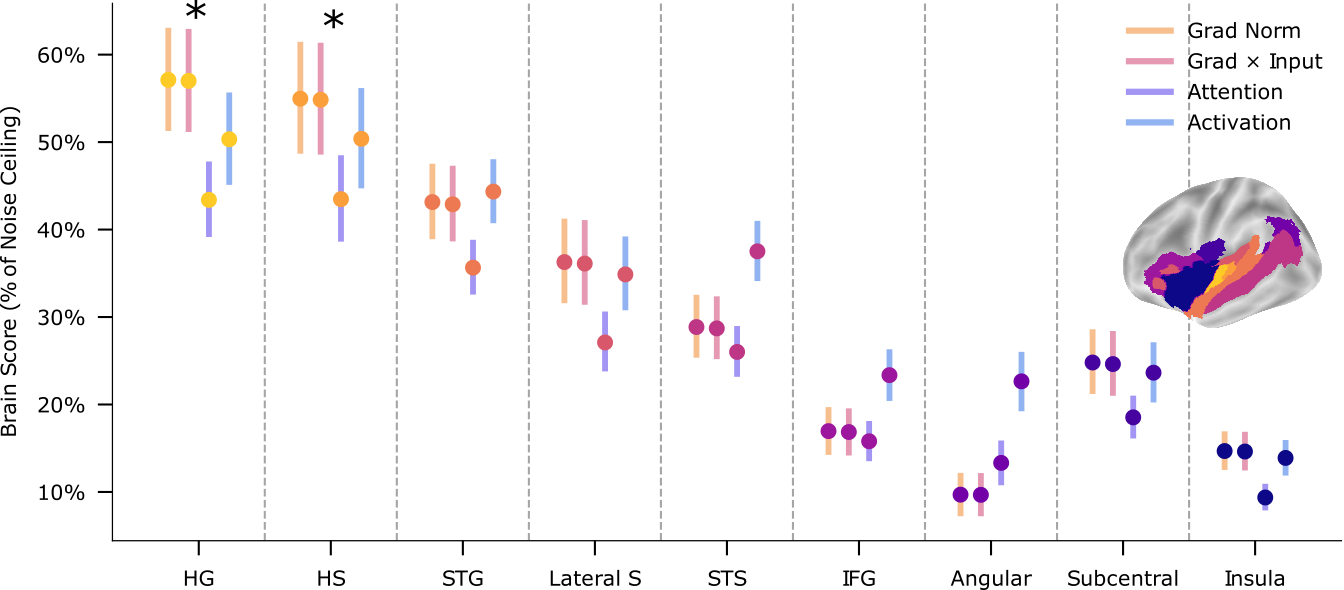}
\caption{
\textbf{Comparing the brain alignment of LLM explanations and internal representations.} 
Dot colors indicate ROIs as shown in the brain map; error-bar colors denote different feature spaces. Asterisks mark ROIs where both attribution methods significantly exceeded activation (FDR-corrected paired bootstrap, $p<0.05$).}
\vspace{-5mm}
\label{fig:Feature_space_comparison}
\end{figure}

\begin{figure*}[!t]
\centering
\includegraphics[width=\linewidth]{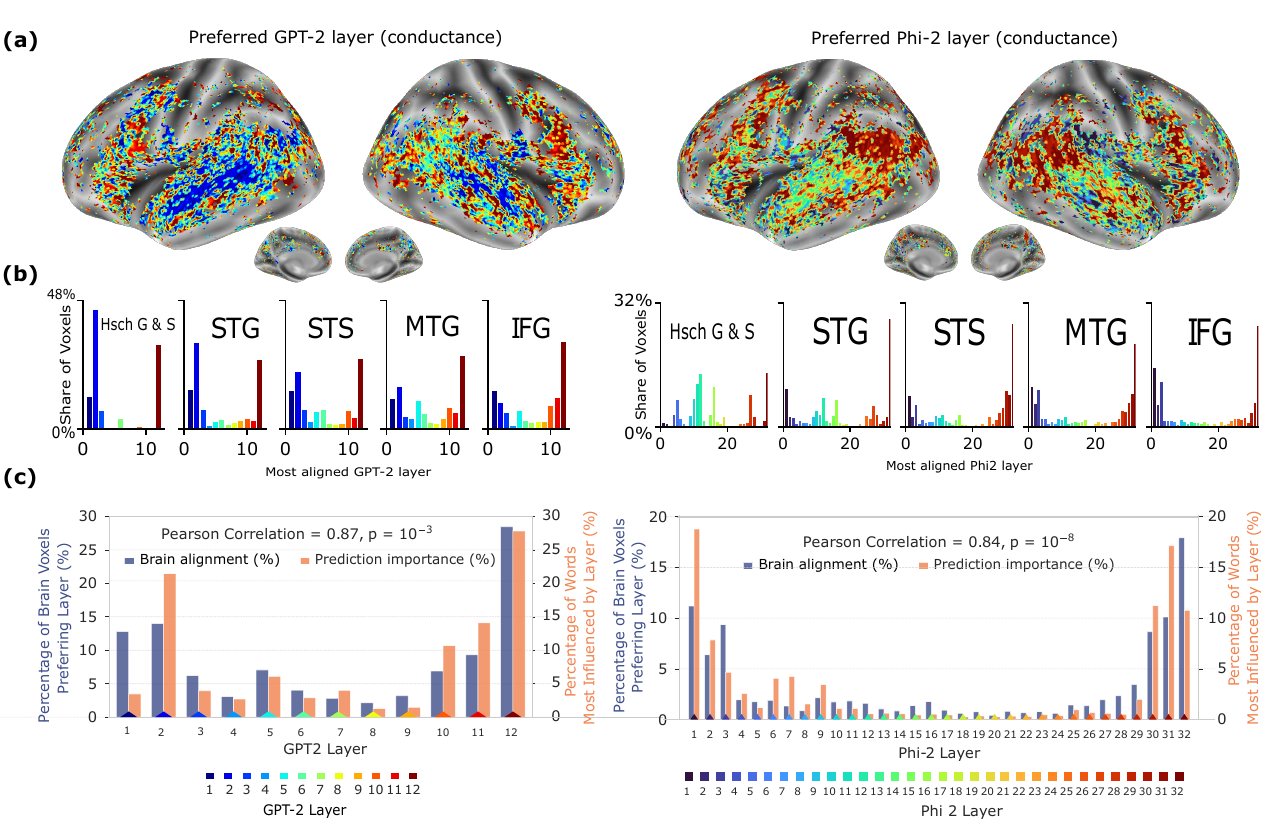}

\caption{
\textbf{Layer preference of conductance explanations across the cortical hierarchy.}
\textbf{(a)} Layer preference maps showing, for each voxel, the model layer whose conductance achieves the highest brain alignment (FDR-corrected $P<0.01$).
\textbf{(b)} Distribution of layer preference across language ROIs. Early layers are preferred in auditory and superior temporal regions, whereas both early and late layers are preferred in higher-order language regions such as MTG and IFG.
\textbf{(c)} Relationship between layer importance (orange; proportion of words for which each layer has maximal conductance) and brain alignment (blue; proportion of voxels preferring each layer). Layers more important for the model's predictions are also preferred by more brain voxels (GPT-2: $r=0.87$, $P=10^{-3}$; Phi-2: $r=0.84$, $P=10^{-8}$). Llama-2 results are shown in Figure~\ref{fig:A_llama2_conductance}.}
\vspace{-5mm}
\label{fig:Layer_alignment}
\end{figure*}

Comparing brain alignment between attribution-based explanations and internal representations, we find that Gradient $\times$ Input and Gradient Norm achieve significantly higher brain alignment than both internal representation baselines in the auditory cortex. By contrast, activations show higher alignment in the inferior frontal gyrus (IFG) and angular gyrus (AG), consistent with these regions' roles in semantic integration \cite{IFG,AG} (Figure~\ref{fig:Feature_space_comparison}; right-hemisphere results in Figure~\ref{fig:A_features_rh}).

This brain alignment pattern holds despite attribution's much lower dimensionality (10 features for explanations versus 768 for activations). To ensure this is not merely an artifact of ridge regression penalizing higher-dimensional spaces, following \citet{caucheteux2023}, we repeated the analysis after reducing activations to 20 principal components using PCA. This dimensionality reduction slightly decreased activation encoding performance, widening the explanations' comparative advantage to four of the nine tested ROIs (Figure~\ref{fig:A_PCA}).
This suggests that the observed advantage of attribution-based explanations is not solely explained by differences in feature dimensionality.

\subsection{Layer-wise Explanations Reveal Distinct Patterns of Brain Alignment}\label{explanations_predict_hierarchy}

We next investigate how conductance -- layer-wise explanations -- aligns with brain activity. After establishing that layer conductance significantly predicts activity in language-related voxels (Figure~\ref{fig:layer_conductance_map}), we compare conductance alignment across model layers per voxel, characterizing each voxel by its \emph{layer preference}: the model layer whose conductance achieves the highest brain alignment. Figure~\ref{fig:Layer_alignment}(a) visualizes these voxel-wise layer preference maps for GPT-2 and Phi-2 (Llama-2 in Figure~\ref{fig:A_llama2_conductance}), while Figure~\ref{fig:Layer_alignment}(b) summarizes the distribution of layer preference in language ROIs.
We find that voxels in primary auditory regions (HG/HS/STG/superior temporal sulcus) prefer early-layer conductance (layers 1--3 for GPT-2; 10--12 for Phi-2; 3--7 for Llama2). In higher-order regions such as MTG and IFG, later layers (10--12 for GPT-2; 28--32 for Phi-2; 30--32 for Llama2) increasingly achieve best alignment alongside early layers, but the last layer's conductance consistently predicts a large share of voxels across every ROI, including auditory cortex (Figure~\ref{fig:Layer_alignment}b).

\paragraph{What layer conductance encodes.}
To understand this pattern, we examine what conductance reveals about how each layer weights words in the input (Appendix~\ref{app:pos_sensitivity}). We find that early layers show word-type sensitivity, while later layers converge toward treating all words equally based only on their position in the sequence. This positional bias increases with depth and peaks at the last layer, whose conductance is concentrated almost entirely on the single most recent word (Appendix~\ref{app:positional_bias}). This near-total positional concentration means the last layer's conductance feature space closely approximates a word-rate signal (see Appendix~\ref{app:word_rate_derivation} for more details).
This characterization is consistent with the spatial pattern of layer-preference shown in Figure~\ref{fig:Layer_alignment}(b): early layers, which exhibit higher word type sensitivity, dominate auditory and language regions, while the last layer, which reflects position rather than meaning, is preferred broadly across the cortex.
However, this interpretation remains categorical. We do not observe a consistent graded relationship between the magnitude of a layer's word-type sensitivity and its brain-alignment score across the full range of layers or across all three models (Appendix~\ref{app:sensitivity_alignment_quant}).

\paragraph{Layer importance predicts brain alignment.}

Beyond \emph{where} each layer's conductance aligns, we ask whether a layer's overall importance to the model's own next-word prediction relates to its alignment with brain activity. We compare each layer's importance (the share of words for which it carries maximal conductance, Section~\ref{sec:conductance}) against the distribution of layer preference across brain voxels (Figure~\ref{fig:Layer_alignment}c).

Both distributions are bimodal, dominated by early and late layers on both the model and brain sides (Figure~\ref{fig:Layer_alignment}c; Appendix~\ref{app:layer_dominance} confirms the model-side pattern is not driven by repetition of specific words across the POS categories). The early layer importance is consistent with the word-type sensitivity established in Appendix~\ref{app:pos_sensitivity}. The late layer importance follows from the same positional mechanism: the last layer becomes the most important layer for words whose conductance was low at earlier layers but who occupied the most recent position in the window, since the last layer concentrates conductance there regardless of word identity. On the brain side, voxels similarly favor early or late layers over the middle of the network; a comparable late-layer-skewed distribution of voxel layer preference has been reported for raw activations \citep{schrimpf2021}, suggesting this shape is not specific to our feature space.
The two distributions are correlated for all three models (GPT-2: $r = 0.87$, $P = 10^{-3}$; Phi-2: $r = 0.84$, $P = 10^{-8}$; Llama2: $r = 0.91$, $P = 10^{-12}$)
and this relationship is robust to excluding the last layer and to restricting to language-network voxels; robustness to excluding both tails of the layer range varies by model (Table A\ref{tab:layer_importance_robustness}).

\section{Discussion}\label{discussion}

We used attribution methods from XAI as an interpretable feature space for studying LLM-brain alignment. Beyond establishing that attribution aligns with brain activity, this interpretability lets us ask what the aligned signal represents -- a question raw alignment scores cannot answer on their own.

Layer conductance made this possible directly: because conductance decomposes into interpretable dimensions (word identity, relative position), we could compare how each layer weighs words by content versus by position, a comparison a high-dimensional activation vector does not support in the same way. This is how we identified the dissociation reported in Section~\ref{explanations_predict_hierarchy}: early layers differentiate words by content, while the last layer collapses toward position alone, explaining why its alignment is broad and non-selective rather than regionally concentrated.

By construction, attribution measures sensitivity to local change in the input -- how much the model's output shifts with a given word -- rather than a model's accumulated internal state, which is what activations summarize. Consistent with this, we find that attribution features themselves, like early-layer conductance, differentiate words by content (Section Appendix~\ref{app:attribution_conductance_characterization}). This offers one plausible account for why attribution's advantage over activations is concentrated in early auditory regions specifically, rather than distributed evenly across the language network: a signal built from local sensitivity to word-level change may be a better match for regions tracking transient input properties than for regions integrating meaning over longer spans, such as IFG and angular gyrus. Because our attribution feature aggregates a word's contribution to predicting upcoming words (Section~\ref{sec:attribution}), we cannot fully rule out that this temporal aggregation, rather than word-level content sensitivity alone, contributes to the regional pattern observed; we therefore note this as an interpretation consistent with our results, not a claim we test directly.

For XAI research, this suggests brain alignment is a useful source of external evidence for what an explanation method captures, provided it is paired with independent characterization of the signal itself, rather than treated as validation on its own. For language neuroscience, attribution offers a concrete, testable operationalization of context-weighting; the regional specificity we observe suggests such comparisons are most informative when paired with an account of what is driving them, particularly when alignment is broad rather than regionally selective.

\clearpage
\section{Limitations}

\textbf{Window length.} Our attribution and conductance feature spaces use a 10-word sliding window, shorter than the context available to the activation baseline (1024 tokens for GPT-2 and Llama-2, 2048 for Phi-2; Section~\ref{sec:baselines}). This window length follows prior work (Section~\ref{sec:attribution}) and is constrained by cost: computing attribution requires a full forward and backward pass per window and layer, substantially more expensive than the forward-only pass used for activation extraction. Whether attribution's relative advantage over activations holds at longer context lengths remains untested.

\textbf{Timing of aggregated attribution scores.} Because a word's attribution feature aggregates its contribution across all windows in which it appears, its feature vector can reflect the model's use of that word in predicting target words up to nine positions later, while being timestamped at the word's own occurrence. This property is shared with deliberate manipulations of future/predictive information in prior brain-alignment work (e.g., \cite{caucheteux2023}, who explicitly test predictive processing by incorporating future context). Whether this aspect of our feature space is similarly informative for modeling predictive processing in the brain, or instead constitutes a confound requiring a causal-only control, is an open question our design does not resolve.

\textbf{Acoustic control at the layer level.} Our acoustic/word-rate control (Section~\ref{sec:acoustic_control}) is applied to the whole-model attribution feature space; we do not repeat this variance-partitioning analysis for each layer's conductance individually. The last layer's approximation to word rate is instead established analytically (Appendix~\ref{app:word_rate_derivation}), showing its feature vector converges to a near-deterministic function of word count per TR. A direct empirical test at each layer would strengthen this claim but was not run given the computational cost of the full acoustic-control pipeline across the network's full depth.

\textbf{Activation layer selection.} We compare attribution against activations from a single layer per model, selected based on prior literature (\citet{caucheteux2023}; \citet{mischler2024contextual}) rather than selecting the empirically best layer for each brain region in our own data. This provides a consistent baseline across analyses, although future work could compare attribution against ROI-specific optimal activation layers.

\textbf{Future Work.} We focus on gradient-based attribution methods and one perturbation-based method (Erasure); more recent techniques designed to improve faithfulness and stability, such as Sequential Integrated Gradients \citep{enguehard2023sequential}, remain untested in this work. 
More broadly, future work could test whether brain alignment systematically relates to explanation quality or faithfulness across attribution methods. If such a relationship exists, brain alignment may provide an additional external criterion for evaluating XAI methods.

\bibliography{refs }


\clearpage
\FloatBarrier 
\appendix
\FloatBarrier
\renewcommand{\thefigure}{A\arabic{figure}} \setcounter{figure}{0}
\renewcommand{\thefigure}{A\arabic{figure}}

\section{Appendix A: Experimental Details}
\subsection{Datasets}
\label{app:datasets}

We used three datasets from the \textit{Narratives} collection \citep{nastase2021narratives}: \textit{Pieman}, \textit{Shapes}, and the paired stories \textit{Slumlord} and \textit{Reach for the Stars One Small Step at a Time}. All datasets consist of fMRI recordings acquired while participants listened to spoken narratives under naturalistic conditions.

The \textit{Pieman} dataset contains recordings from 86 participants (18--45 years, mean age $22.5\pm4.3$ years; 45 reported female) listening to a 7:02-minute story (282 TRs, 957 words). The \textit{Shapes} dataset contains recordings from 58 participants (18--35 years, mean age $23.0\pm4.5$ years; 42 reported female). This dataset includes two narrative descriptions of the same animation (\textit{shapesphysical} and \textit{shapessocial}); We used the \textit{shapessocial} condition (6:45 minutes, 270 TRs, 910 words). The \textit{Slumlordreach} dataset comprises recordings from 18 participants (18--27 years, mean age $21.0\pm2.3$ years; 8 reported female) who listened to two consecutive stories during the same scanning session. The \textit{Slumlord} story lasts 15:03 minutes (602 TRs, 2,715 words), while \textit{Reach for the Stars One Small Step at a Time} lasts 13:45 minutes (550 TRs, 2,629 words).

We used the preprocessed cortical-surface data released as part of the Narratives collection. The original dataset was preprocessed using the fMRIPrep pipeline, including susceptibility-distortion correction, slice-timing correction, spatial normalization to the fsaverage cortical surface, and projection of functional data onto the cortical surface. We used the unsmoothed version of the data, i.e., no spatial smoothing had been applied to the released data, and we performed no additional preprocessing before feature extraction or encoding analyses. Functional images were sampled at a repetition time (TR) of 1.5 s, and no temporal filtering was applied.

The Narratives release also provides recommended recording exclusions based on measures of neural entrainment in the left early auditory cortex.
Specifically, scans were excluded if (1) the leave-one-out inter-subject correlation (ISC) was below 0.10 or (2) the peak ISC occurred at a temporal lag exceeding ±3 TRs (4.5 s) from the stimulus.
Following these recommendations, our analyses included 150 recordings from 147 unique participants; three participants contributed recordings to more than one dataset in separate scanning sessions.

\paragraph{Brain Parcellation}\label{methods_brain_rep_parcel}
All encoding analyses were performed at the voxel level. For ROI-level analyses and visualization (Figures \ref{fig:Attribution_predict} b and \ref{fig:Feature_space_comparison}), voxel-wise brain scores were summarized within anatomically defined regions of interest (ROIs) using the Destrieux atlas \citep{Destrieux2010}, which parcellates each hemisphere into 75 cortical regions. The ROI abbreviations used throughout the paper are listed in Table~\ref{tab:roi_abbreviations}.

\begin{table}[h]
\centering
\caption{ROI abbreviations used throughout the paper, based on a subdivision of the
Destrieux atlas \citep{Destrieux2010}. The 9-region set is used for the
main ROI-level comparisons (Figures~\ref{fig:Attribution_predict}b,
\ref{fig:Feature_space_comparison}); the reduced 5-region set is used for the
layer-conductance analysis (Figure~\ref{fig:Layer_alignment}).}
\begin{tabular}{ll}
\hline
\textbf{Abbreviation} & \textbf{Full Name} \\
\hline
\multicolumn{2}{l}{\textit{9-region set}} \\
HG          & Heschl's Gyrus \\
HS          & Heschl's Sulcus \\
STG         & Superior Temporal Gyrus \\
STS         & Superior Temporal Sulcus \\
Lateral S   & Lateral Sulcus (Sylvian Fissure) \\
IFG         & Inferior Frontal Gyrus \\
Angular     & Angular Gyrus \\
Subcentral  & Subcentral Gyrus and Sulci \\
Insula      & Insula \\
\hline
\multicolumn{2}{l}{\textit{5-region set (layer-conductance analysis)}} \\
Hsch G \& S & Heschl's Gyrus and Sulcus \\
STG         & Superior Temporal Gyrus \\
STS         & Superior Temporal Sulcus \\
MTG         & Middle Temporal Gyrus \\
IFG         & Inferior Frontal Gyrus \\
\hline
\end{tabular}
\label{tab:roi_abbreviations}
\end{table}

\subsection{Target Selection and Baselines for Attribution Computation}
\label{app:target_conditioning}
For all attribution methods, we computed attribution with respect to the actual next word in the story, rather than the model's predicted distribution over possible next tokens, so that explanations were grounded in the linguistic input heard by the participants. Concretely, attribution is computed with respect to the raw logit of this target token for Gradient Norm, Gradient $\times$ Input, and Integrated Gradients, and with respect to its softmax probability for Erasure.
When the target word was tokenized into multiple subword tokens, attribution was computed with respect to the first subword token. For Llama-2 and Phi-2, if the first subword decoded to an empty string (a tokenizer artifact caused by leading whitespace), we instead used the second subword token. 
The baseline for Integrated Gradients was the default baseline of Captum's \texttt{LLMGradientAttribution} implementation.

For Layer Conductance, attribution was computed with respect to the actual next word in the story by conditioning on the first token of the target word when it consisted of multiple subword tokens. The baseline input consisted of a reference sequence in which all input token IDs were replaced with 0, and the corresponding input embeddings were used as the conductance baseline.

\subsection{Acoustic Feature Extraction}
\label{app:acoustic_details}

This section describes the implementation details of the acoustic baseline used in the variance-partitioning analysis.

\paragraph{Mel-Power Spectrogram}
We computed a 128-channel mel-power spectrogram from the audio waveform using a Fast Fourier Transform ($n_{\text{fft}}=2048$) with a 10\,ms hop length to capture the spectral energy distribution across frequency bands.

\paragraph{Root Mean Square (RMS) Energy}
We computed the root mean square (RMS) energy over the same 10\,ms frames to capture continuous variations in speech intensity.

\paragraph{Word Rate}
To account for temporal speech rhythm, we computed the number of word onsets occurring within each fMRI repetition time (TR = 1.5\,s) using the word-level timing annotations provided with the Narratives dataset.

Mel-spectrogram and RMS features were downsampled to the fMRI acquisition rate by averaging all 10\,ms frames within each TR interval (150 frames/TR). The resulting 129 acoustic features were concatenated with the TR-level word-rate feature, yielding a 130-dimensional acoustic baseline used in the variance-partitioning analysis.

\subsection{Brain Encoding Details}
\label{app:encoding_details}

This section provides the implementation details of the voxel-wise encoding framework used throughout the paper.

We first aligned all feature spaces to the temporal resolution of the fMRI acquisition (TR = 1.5 s). For the word-level feature spaces (activations, attention, attribution methods, and layer conductance), feature vectors corresponding to all words presented within the same TR were summed to model the cumulative neural response \citep{Huth2016,caucheteux2023}. For the acoustic baseline, mel-spectrogram and RMS features were first extracted at 10 ms resolution and then averaged across all frames within each TR interval (150 frames/TR), while word rate was computed directly at the TR level.

Following \citet{Huth2016}, we modeled the delayed BOLD response using a finite impulse response (FIR) model by concatenating feature vectors from the current and five preceding TRs (0--7.5 s). Each feature space was then processed using an encoding pipeline implemented in \texttt{scikit-learn}, consisting of (1) feature standardization using \texttt{StandardScaler}, (2) principal component analysis (PCA), retaining the first 20 components for the attention feature space only because of its substantially higher dimensionality, and (3) $\ell_2$-regularized linear regression (\texttt{RidgeCV}). Results obtained when applying PCA to the activation feature space are reported in Figure~\ref{fig:A_PCA}.

The regularization hyperparameter ($\alpha$) of \texttt{RidgeCV} was optimized independently for each voxel within every training fold using leave-one-out cross-validation over ten logarithmically spaced values between $10^{-1}$ and $10^8$. Model performance was evaluated using five-fold cross-validation, where each fold corresponded to a temporally contiguous segment of the story. The final \emph{brain score} was computed as the Pearson correlation between predicted and observed BOLD responses on each held-out fold and averaged across the five test folds \citep{yamins2014}.
\begin{figure*}[!tb]
    \centering
    \includegraphics[width=0.6\textwidth]{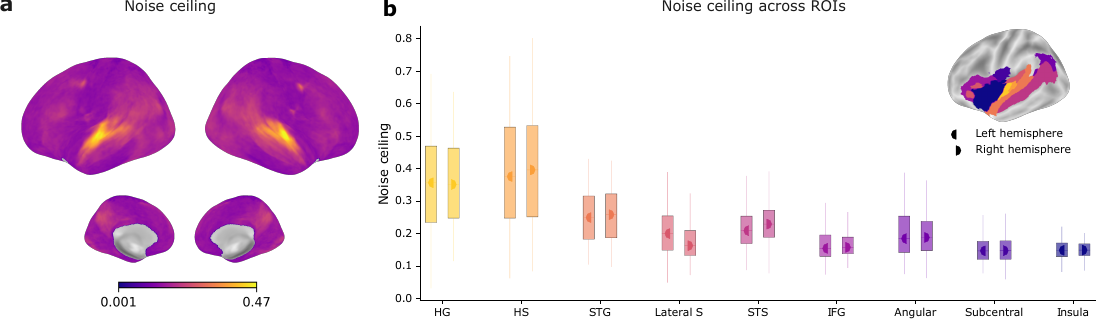}
    \caption{\textbf{Noise Ceiling Estimation.} \textbf{(a)} Noise ceiling values for each voxel \textbf{(b)} Noise ceiling values for each ROI. Boxplots represent the distribution of averaged noise ceiling values across individuals. \label{fig:S_noise_ceiling}}
    \vspace{-5mm}
\end{figure*}

\subsection{Statistical Details}
\label{app:stats}

\paragraph{Whole-cortex significance maps (e.g., Figure~\ref{fig:Attribution_predict} (a))} At each
voxel, we test whether the average score is significantly different from zero using a Wilcoxon signed-rank test across subjects.
$P$-values are FDR corrected across all voxels in the map ($q<0.05$).

\paragraph{ROI-level comparisons.} For each ROI, we compute one score per subject by averaging across the voxels within that ROI; where a
figure aggregates across language models, we additionally average across the three models per subject before the ROI average. We then
use a bootstrap hypothesis test: subjects are resampled with replacement (10,000 iterations) to build a distribution of the mean,
from which we compute a 95\% confidence interval and a $p$-value.
Tests are one-sided when checking whether a score is significantly above zero, and two-sided when comparing two methods, models, or
feature spaces (using the subject-level paired difference). $P$-values are FDR corrected within each figure's family of
comparisons (e.g., across ROIs, or across method pairs).

\paragraph{Layer preference maps (Figure~\ref{fig:Layer_alignment} (a)} At each voxel, we first
identify which layers significantly predict that voxel using a one-sided Wilcoxon signed-rank test, FDR corrected at a stricter threshold ($q<0.01$) than used elsewhere, since selecting a best layer involves comparing across all layers per voxel. Among only the layers that significantly predict a given voxel, we then compare their mean brain scores and assign the best-predicting layer, provided its score also exceeds a minimum magnitude threshold.

\subsection{Noise Ceiling}\label{app:methods_noise_ceiling}
The noise ceiling provides an estimate of the maximum explainable prediction accuracy, serving as an upper bound for evaluating encoding models. We estimated a voxel-wise noise ceiling using inter-subject correlation (ISC) \cite{Nastase_isc}, a standard measure of the reliability of stimulus-evoked responses that is also used by the Narratives dataset for participant quality assessment. For noise-ceiling estimation, we followed the cross-validated ISC procedure of \citet{kumar2024shared}.

We implemented the same five-fold framework  used for brain encoding, where each fold represents a temporally contiguous segment of the story stimulus. For each of the five folds, we computed the Pearson correlation between an individual’s voxel responses and the average responses of the other participants during that same fMRI time-series fold. This procedure resulted in five ISC values per voxel for each recording. The final noise ceiling for each voxel was obtained by averaging these correlations across all folds and participants. 

Following \citet{kumar2024shared}, the group average includes the target participant to account for autocorrelation, resulting in a more conservative (higher) estimate of the noise ceiling.
Voxel-wise and ROI-level noise ceiling estimates are visualized in Figure~\ref{fig:S_noise_ceiling}, showing the spatial distribution of response reliability across the cortex and the corresponding values within language-related ROIs.
In the main text, ROI-level brain scores are reported as a percentage of this upper bound. For variance-partitioning analyses, unique variance is normalized by the squared ROI noise ceiling to express the fraction of explainable variance.

\section{Appendix B: Additional Results}
\subsection{Attribution-Based Brain Alignment}
\label{app:additional_alignment}

\begin{figure}[!b]
    \centering
    \includegraphics[width=0.4\textwidth]{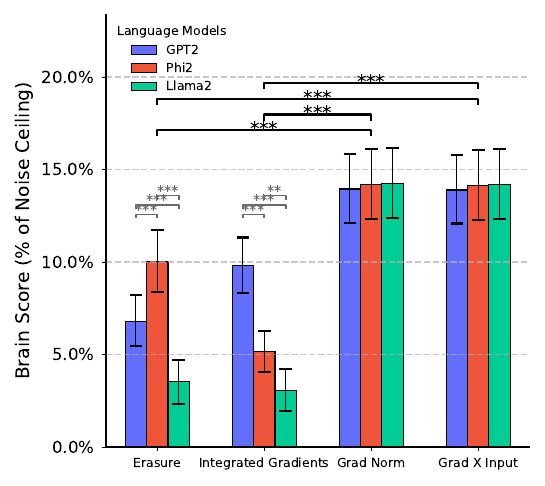}
    \vspace{-3mm}
    \caption{\textbf{Attribution method comparison across LLMs.} Bars show mean
    normalized brain scores; error bars are 95\% bootstrap CIs. Black brackets:
    significant attribution-method comparisons; gray brackets: significant LLM
    comparisons within a method (FDR-corrected paired bootstrap; *$p<0.05$,
    **$p<0.01$, ***$p<0.001$). Gradient Norm and Gradient$\times$Input differed
    significantly ($p_\mathrm{FDR}=0.035$); shown separately in
    Figure~\ref{fig:A_model_grad_norm_vs_grad_x_input}. 
    See Appendix~\ref{app:stats} for test details. \label{fig:A_attribution_comparison}}
\end{figure}

\begin{figure*}[h]
    \centering
    \includegraphics[width=1\textwidth]{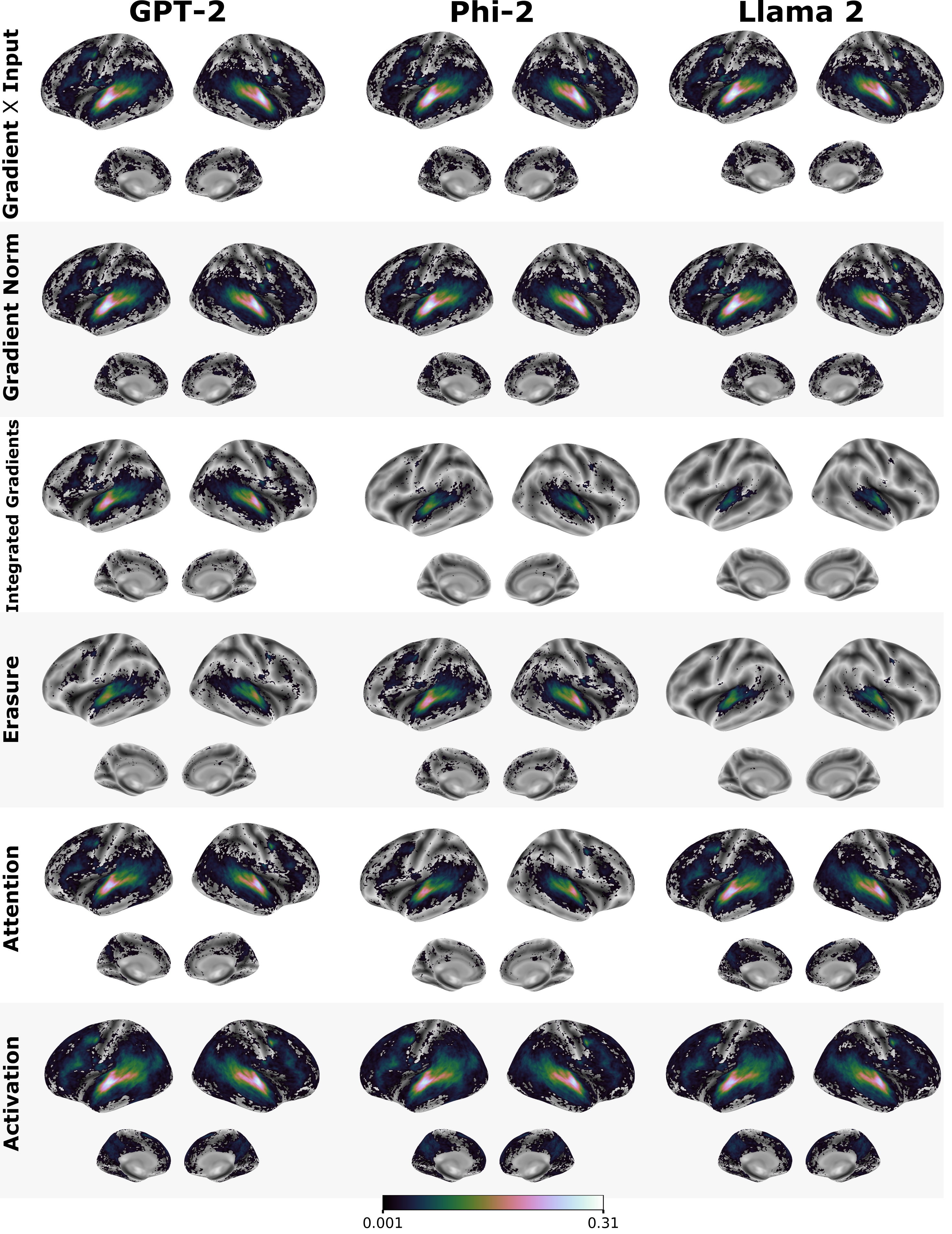}
    \caption{\textbf{Attribution and internal-representation brain maps across LLMs.}
    Only significant voxels are shown (FDR-corrected, $p<0.05$); scores are averaged
    across participants. \label{fig:A_brain_maps_all_features}}
\end{figure*}

\begin{figure*}[!t]
    \centering
    \includegraphics[width=1\textwidth]{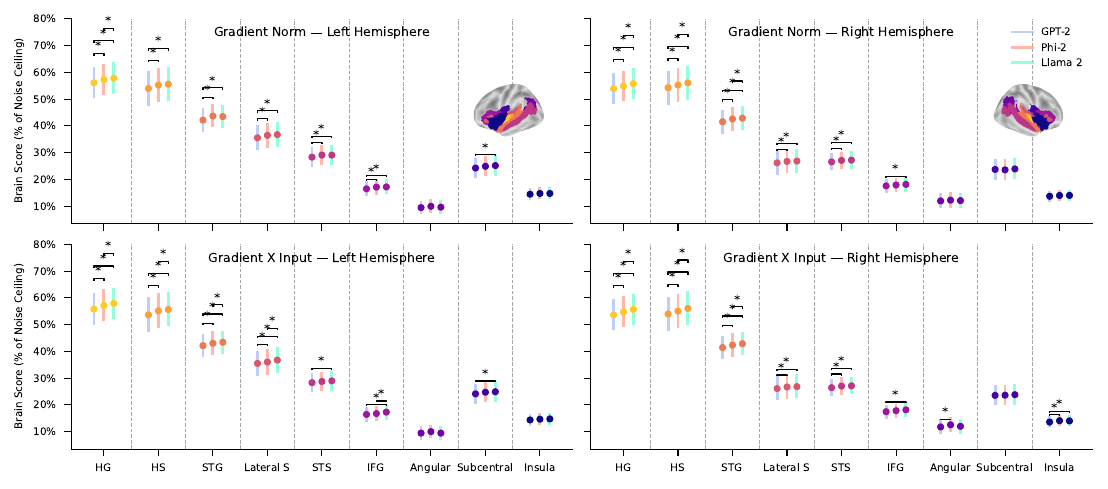}
    \caption{\textbf{Grad Norm vs.\ Grad$\times$Input across LLMs and ROIs.}
    Dot color indicates ROI (matching the inset brain map); error-bar color
    indicates LLM. Asterisks denote significant model comparisons
    ($p_\mathrm{FDR}<0.05$). \label{fig:A_model_grad_norm_vs_grad_x_input}}
    \vspace{-5mm}
\end{figure*}

\begin{figure*}[!t]
    \centering
    \includegraphics[width=1\textwidth]{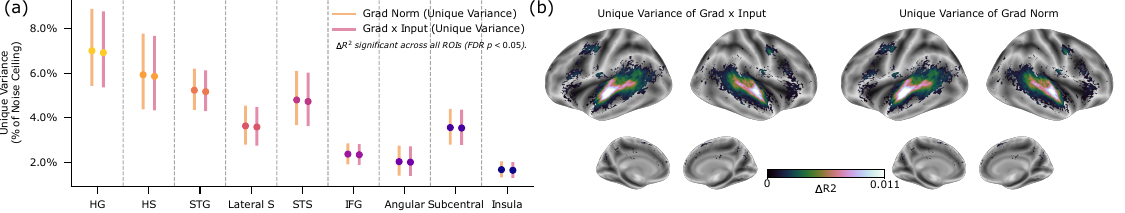}
    \caption{\textbf{Unique variance, right hemisphere.} \textbf{(a)} Unique
    variance per ROI (Figure~\ref{fig:Attribution_predict}b). \textbf{(b)} Unique variance
    per voxel; only voxels with significant $\Delta R^2$ are colored
    ($p_\mathrm{FDR}<0.05$). \label{fig:A_unique_var_rh}}
    \vspace{-5mm}
\end{figure*}

This section provides supplementary analyses for Section~\ref{sec:attribution_predict}. Figure~\ref{fig:A_attribution_comparison} compares brain alignment across attribution methods. Figures~\ref{fig:A_brain_maps_all_features} and \ref{fig:A_model_grad_norm_vs_grad_x_input} present voxel-wise brain-alignment maps and model comparisons for all tested feature spaces. Figure~\ref{fig:A_unique_var_rh} reports the corresponding right-hemisphere and voxel-wise variance-partitioning results.
\FloatBarrier

\vspace{-24mm}


\begin{figure*}[b]
    \centering
    \includegraphics[width=1\textwidth]{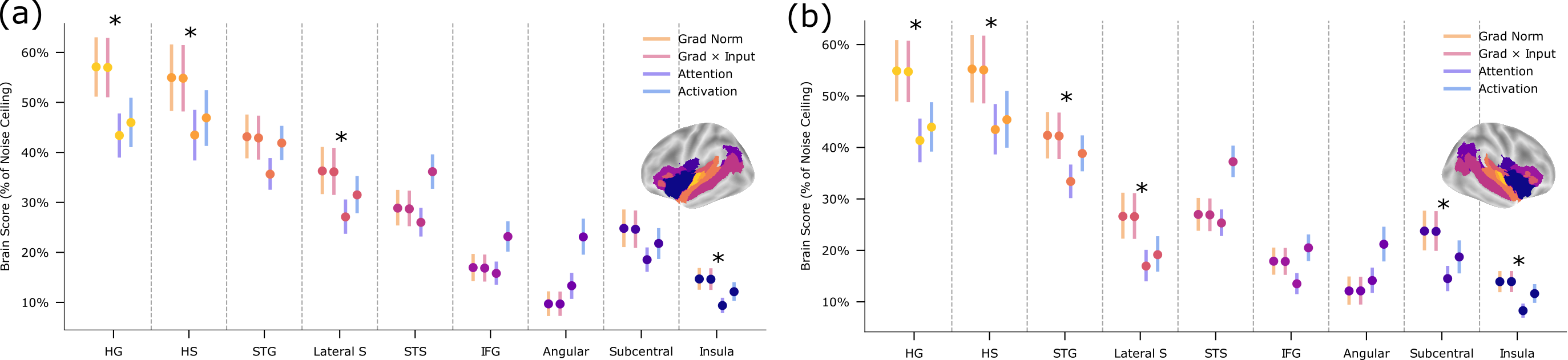}
    \caption{
    \textbf{Attribution vs.\ PCA-matched internal representations.}
    Same as Figure~\ref{fig:Feature_space_comparison}, but with the activation feature space reduced to 20 principal components using PCA. Dot color indicates ROI; line color indicates feature space. \textbf{(a)} Left hemisphere. \textbf{(b)} Right hemisphere.
    \label{fig:A_PCA}
}
\vspace{-5mm}
\end{figure*}
\FloatBarrier
\subsection{Comparison with Internal Representations}
\begin{figure}[htbp]
    \centering
    \includegraphics[width=0.5\textwidth]{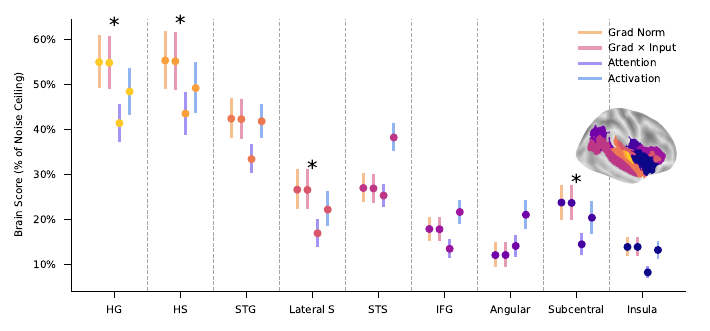}
    \caption{\textbf{Attribution vs.\ internal representations, right hemisphere.}
    Same as Figure~\ref{fig:Feature_space_comparison}, 
    shown for right-hemisphere ROIs. Dot color indicates ROI; line color
    indicates feature space. \label{fig:A_features_rh}}
\end{figure}

This section provides additional analyses supporting the comparison between attribution-based explanations and internal representations in Section~\ref{explanaitions_outperform}. Figure~\ref{fig:A_features_rh} reports the corresponding results for the right hemisphere. Figure~\ref{fig:A_PCA} repeats the comparison after reducing the activation feature space to 20 principal components as a dimensionality control.
\FloatBarrier

\subsection{Layer-wise Brain Alignment}
\label{app:layer_alignment}

\begin{figure}[!t]
    \centering
    \includegraphics[width=1\linewidth]{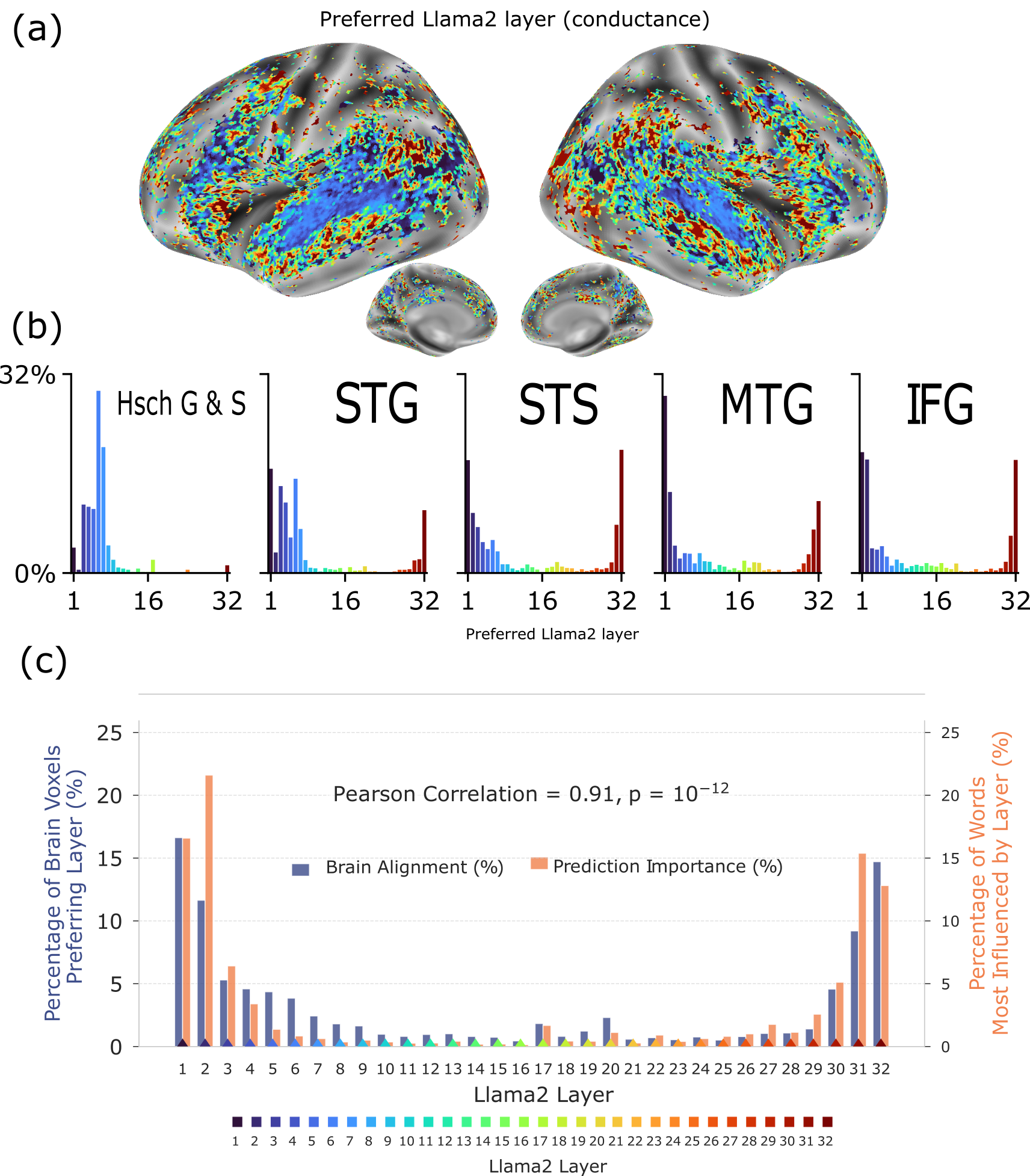}
\caption{
\textbf{Layer preference and layer importance for Llama-2 conductance explanations.}
Same analysis as Figure~\ref{fig:Layer_alignment}, shown for Llama-2.
\textbf{(a)} Layer preference map showing, for each voxel, the model layer whose conductance achieves the highest brain alignment (FDR-corrected $P<0.01$).
\textbf{(b)} Distribution of layer preference across language ROIs.
\textbf{(c)} Relationship between layer importance (orange; proportion of words for which each layer has maximal conductance) and layer preference (blue; proportion of voxels preferring each layer). Layers more important for the model's predictions are also preferred by more brain voxels ($r=0.91$, $P=10^{-12}$).
}

    \label{fig:A_llama2_conductance}
\end{figure}

\begin{figure*}[!t]
    \centering
    \includegraphics[width=0.8\linewidth]{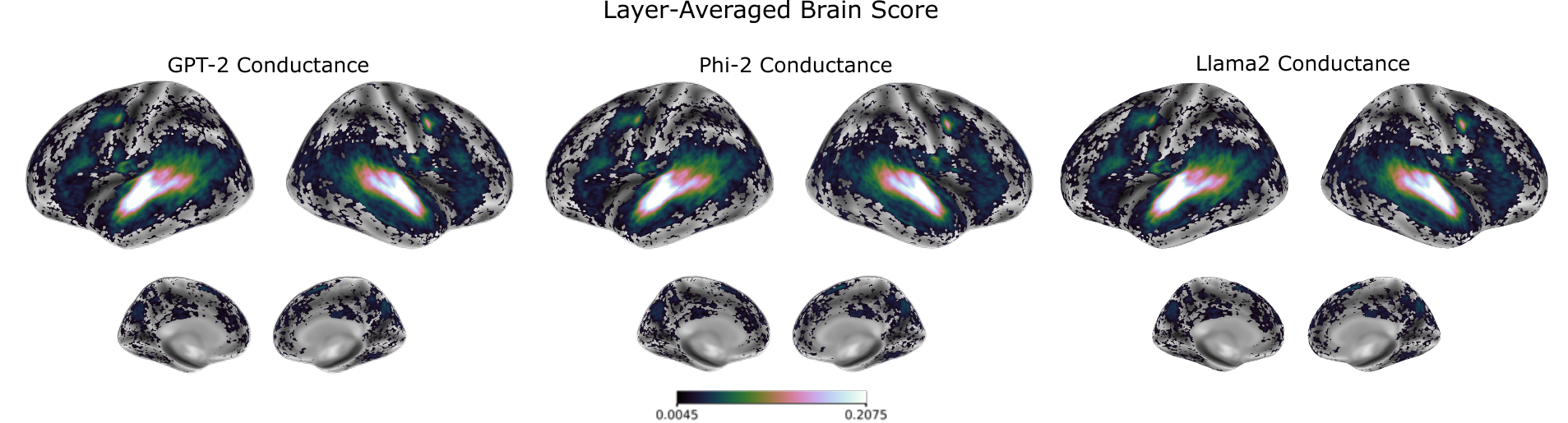}
    \caption{\textbf{Layer conductance predicts a broad range of cortical voxels.}
    Voxels shown are significantly predicted by conductance averaged across all
    layers (FDR-corrected Wilcoxon signed-rank test, $p<0.05$).
    \label{fig:layer_conductance_map}}
\end{figure*}

This section provides additional analyses supporting Section~\ref{explanations_predict_hierarchy}. Figure~\ref{fig:layer_conductance_map} presents whole-brain maps of layer conductance alignment for each model. Figure~\ref{fig:A_llama2_conductance} extends the layer-preference analysis of Figure~\ref{fig:Layer_alignment} to Llama-2. Table~\ref{tab:layer_importance_robustness} reports robustness analyses of the relationship between layer importance and layer preference. 
\begin{table*}[!t]
\centering
\small
\caption{\textbf{Robustness of the correlation between layer importance and layer preference.} Pearson correlation coefficients under different analysis settings.}
\label{tab:layer_importance_robustness}
\begin{tabular}{lcccc}
\toprule
\textbf{Model} &
\textbf{Full} &
\textbf{Excl. last layer} &
\textbf{Excl. first/last 3 layers} &
\textbf{Language voxels} \\
\midrule
GPT-2 &
$r=0.873$, $p=10^{-3}$ &
$r=0.725$, $p=10^{-1}$ &
$r=0.839$, $p=10^{-1}$ &
$r=0.861$, $p=10^{-3}$ \\

Phi-2 &
$r=0.836$, $p=10^{-8}$ &
$r=0.905$, $p=10^{-11}$ &
$r=0.448$, $p=10^{-1}$ &
$r=0.699$, $p=10^{-5}$ \\

Llama-2 &
$r=0.914$, $p=10^{-12}$ &
$r=0.920$, $p=10^{-12}$ &
$r=0.573$, $p=10^{-2}$ &
$r=0.845$, $p=10^{-8}$ \\
\bottomrule
\end{tabular}
\end{table*}

\begin{figure*}[!t]
    \centering
    \includegraphics[width=0.7\linewidth]{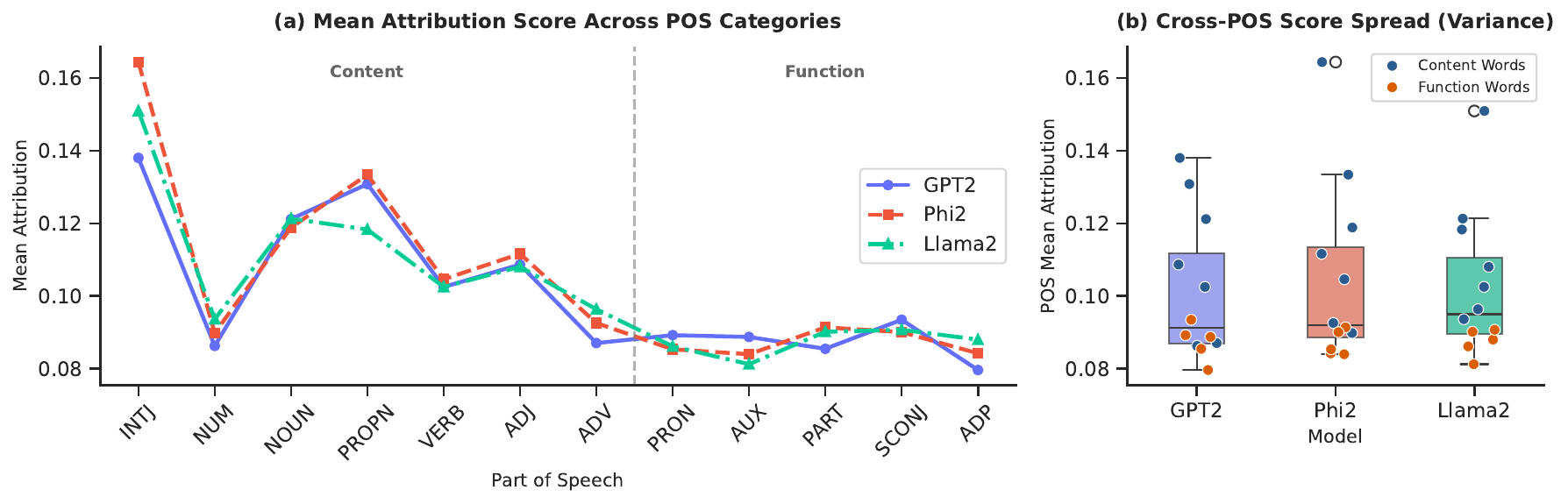}
    \vspace{-3mm}
    \caption{\textbf{Attribution by part-of-speech.} \textbf{(a)} Mean attribution score per POS category, averaged over window position and pooled across stories, shown separately for each model. Content-word categories (left of dashed line) span a wide range, from interjections (highest) to adverbs and numerals (near function-word levels); function-word categories (right) cluster tightly at the low end across all three models. \textbf{(b)} Distribution of per-POS mean scores, grouped as content vs. function words. Content words show a wider score range and higher median than function words for all three models, indicating that attribution differentiates word type rather than assigning uniform importance.}
    \label{fig:word-sensitivity_att}
    \vspace{-8mm}
\end{figure*}
\section{Appendix C: Characterizing Attribution and Layer Conductance}
\label{app:attribution_conductance_characterization}

Because we use overlapping sliding windows, each word passes through up to 10 windows and occupies a different relative position in each one. We store the importance score for each occurrence, giving both our attribution and layer-conductance feature spaces the shape $(\text{words} \times \text{positions})$, where every dimension has a direct interpretation: how much importance a word received when it occupied a given relative position to the prediction. This interpretability lets us ask, how each feature space weights word identity and word position---a question a high-dimensional activation vector does not support.

\subsection{Attribution Feature Space}

\paragraph{Word-type sensitivity in attribution.}
We first ask whether attribution scores track word identity or only word position. Averaging across the position dimension gives an importance score per word, independent of where it appeared in the input; grouping these by part-of-speech (POS, tagged using the \texttt{spaCy} transformer-based POS tagging model \texttt{en\_core\_web\_trf} \cite{honnibal2020spacy}
, pooled across all three stories) shows how each model weights different word types (Figure~\ref{fig:word-sensitivity_att}). All three models converge on a similar ranking: interjections receive the highest attribution, adpositions the lowest, with the remaining content categories (nouns, proper nouns, adjectives) spread across a wide middle range while function-word categories cluster tightly near the low end. This spread is reflected directly in the variance: content words show a substantially wider range of scores across POS categories than function words, which stay narrowly clustered regardless of model (Figure~\ref{fig:word-sensitivity_att}, panel~b).

\begin{figure}
    \centering
    \includegraphics[width=1\linewidth]{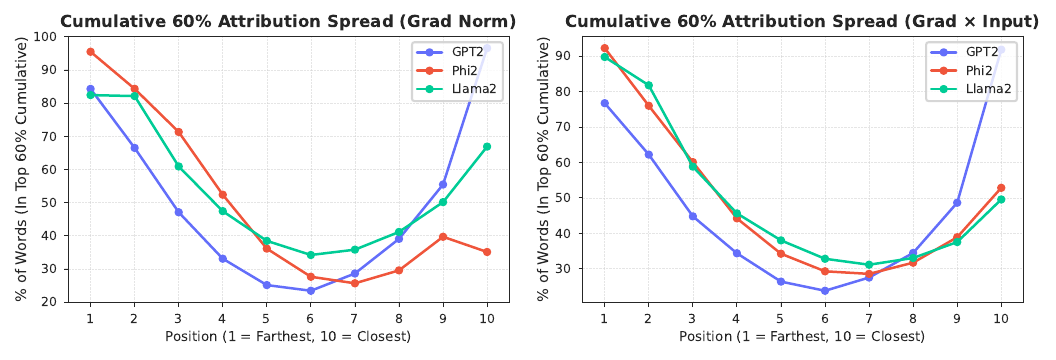}
    \vspace{-1mm}
    \caption{\textbf{Positional concentration of top-attributed words.} For each word, we compute the smallest set of window positions whose cumulative (normalized) attribution mass reaches 60\%, following \citet{proietti2025finegrainedanalysisbrainllmalignment}, then report the percentage of words for which each position falls in that set. \textbf{Left:} Gradient Norm. \textbf{Right:} Gradient $\times$ Input. Both methods show a bimodal profile, with elevated concentration at the window's extremes (positions 1 and 10) relative to the middle. The primacy peak (position 1) matches or exceeds the recency peak (position 10) for Phi-2 and Llama-2; GPT-2 shows a comparatively symmetric profile.}
    \label{fig:position-bias_Att}
\end{figure}

\paragraph{Positional structure in attribution.}
To check for a positional bias independent of word identity, we follow \citet{proietti2025finegrainedanalysisbrainllmalignment} and examine which positions contribute to the top-60\% cumulative attribution mass. Both Gradient Norm and Gradient$\times$Input show a pronounced bimodal profile, with most top-attributed words falling at the two extremes of the window and a marked drop in the middle (Figure~\ref{fig:position-bias_Att}). As in their finding that next-word prediction attribution shows a primacy bias at or exceeding recency bias, the primacy peak matches or exceeds the recency peak for Phi-2 and Llama-2, while GPT-2 shows a more symmetric profile. Taken together, these two analyses show that attribution carries a real positional skew, but this does not come at the expense of word-identity sensitivity: middle positions still receive substantial mass, and word-type differentiation holds throughout.

\subsection{Layer Conductance Feature Space}
\begin{figure}[!htp]
    \centering
    \includegraphics[width=1\linewidth]{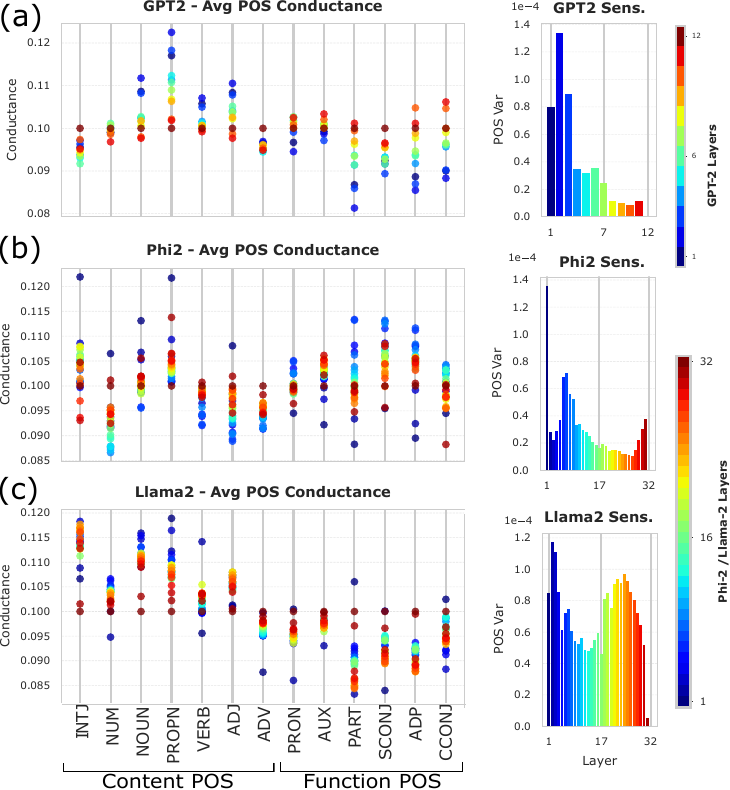}
    \caption{
    \textbf{Layer-wise conductance across part-of-speech (POS) categories.}
    \textbf{(a)} GPT-2, \textbf{(b)} Phi-2, and \textbf{(c)} Llama-2. In each panel, the left plot shows the average layer conductance for each POS category, averaged across all words. Earlier layers assign higher conductance to content words than to function words, whereas this distinction gradually diminishes with depth, converging toward similar conductance values in the final layers. The right plot summarizes each layer's \emph{word-type sensitivity}, quantified as the variance of average conductance across POS categories. GPT-2 exhibits an early peak followed by a nearly monotonic decline in sensitivity. Phi-2 and Llama-2 additionally show a secondary increase in middle-to-late layers, most pronounced for Llama-2 (layers 18--26).
    \label{fig:pos_averaged}}
\end{figure}

\paragraph{Word-type sensitivity in layer conductance.}\label{app:pos_sensitivity}
We apply the same characterization to conductance, now comparing across layers rather than across models. Averaging conductance across positions and then by POS shows that the last layer assigns a near-uniform score of $0.1$ to every word type---the baseline expected under our 10-position window ($1/P$) if conductance is concentrated on a single position regardless of word identity. Early layers deviate substantially from this baseline, with consistently higher across-POS variance, indicating that word-type differentiation is strongest early and fades with depth though not strictly monotonically for all models (Figure~\ref{fig:pos_averaged}).

\begin{figure}[!t]
    \centering
    \includegraphics[width=1\linewidth]{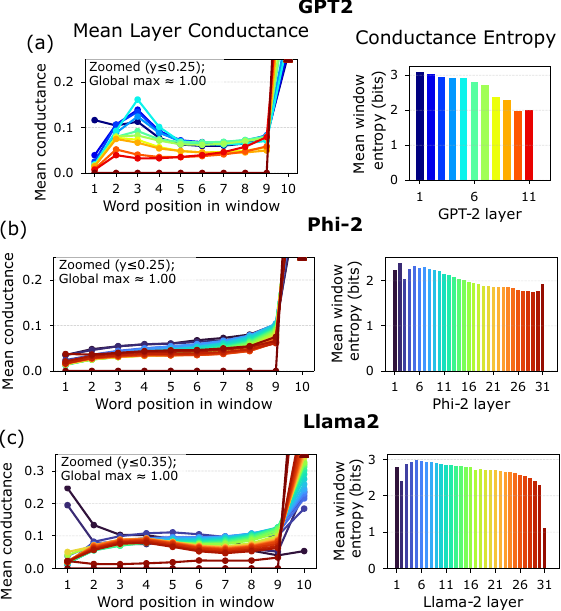}
    \caption{\textbf{ Layer conductance concentrates on recent positions with depth.}
For each model, (left) mean conductance by relative window position, averaged across words, for early vs. late layers; (right) entropy of raw per-window conductance scores by layer depth, computed prior to word-level aggregation. Entropy decreasing with depth confirms that positional concentration in the aggregated feature space reflects genuine layer behavior rather than an artifact of feature construction. \textbf{(a)} GPT-2. \textbf{(b)} Phi-2. \textbf{(c)} Llama2. \label{fig:entropy}}
\end{figure}

\paragraph{Positional structure across layers.}\label{app:positional_bias}
Having found that early layers differentiate conductance by word type while later layers increasingly do not, we next asked whether this loss of differentiation reflects a shift toward positional bias -- that is, whether later layers simply concentrate conductance on specific positions in the window, regardless of word identity.
Following \citet{rahimi2026layerwisepositionalbiasshortcontext}, we average conductance across words to obtain, for each layer, a profile of how conductance is distributed across relative positions; consistent with their findings, later layers concentrate increasingly on the most recent position (Appendix Figure~\ref{fig:entropy}). We verify this is not an artifact of our word-level aggregation by measuring entropy directly on each window's raw conductance scores prior to any aggregation: entropy decreases with depth, confirming that later layers genuinely concentrate conductance on fewer words within a window, rather than this pattern emerging from how we construct the feature space. 
This concentration may partly reflect architectural constraints rather than a learned shift from content to position: in a decoder-only transformer, the next-token prediction head reads only the final hidden state at the last sequence position, with no subsequent attention layer through which earlier positions' representations could be read out directly. We therefore treat this as a diagnostic characterization of what the last layer's conductance reflects, rather than evidence that the model has learned to specialize this way.

\paragraph{Derivation: Last-Layer Conductance as a Word-Rate Proxy}
\label{app:word_rate_derivation}
Because the last layer assigns all conductance to the token immediately preceding the prediction, with all other tokens in the window receiving near-zero conductance, each word's conductance vector across the 10 relative positions approaches $(0,0,0,0,0,0,0,0,0,1)$: full conductance when the word occupies the most recent position, and zero otherwise. When this feature space is downsampled to align with the fMRI sampling rate, each TR's aggregated vector reduces to $(0,0,0,0,0,0,0,0,0,n)$, where $n$ is the number of words in that TR -- a direct proxy for local word rate.

\paragraph{Does Layer Sensitivity or Positional Bias Predict Layer Preference Across ROIs?}
\label{app:sensitivity_alignment_quant}

\begin{figure}[!htp]
    \centering
    \includegraphics[width=1\linewidth]{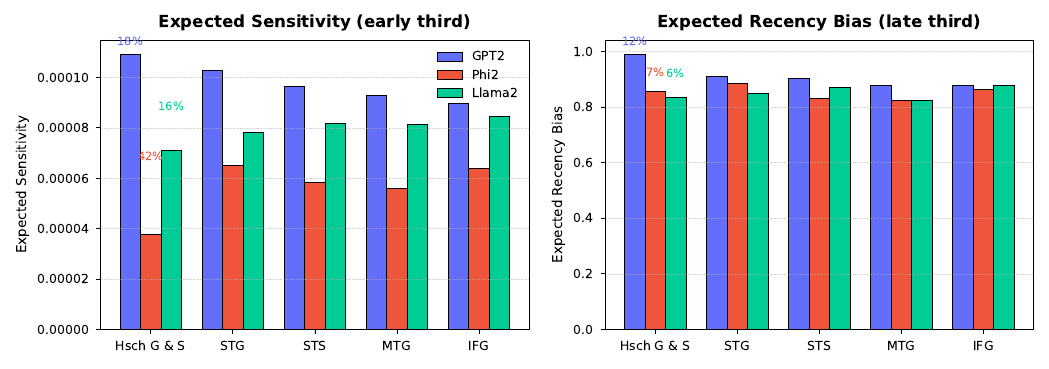}
    \caption{\textbf{Expected layer sensitivity and recency bias per ROI,
    under each ROI's layer-preference distribution.} Left: expected
    word-sensitivity, restricted to the early third of layers. Right:
    expected recency bias, restricted to the late third of layers.
    Percentages above each model's bars indicate the relative range
    (max--min)/max across ROIs. Sensitivity shows a consistent
    ROI-dependent gradient only for GPT-2; bias shows a narrow,
    comparable range across ROIs for all three models.}
    \label{fig:expected_sens_bias}
\end{figure}

Beyond the categorical characterization above, we asked whether the
\emph{magnitude} of a layer's word-sensitivity or positional bias predicts,
in a graded way, how strongly that layer's conductance aligns with the
brain across regions. For each ROI, we computed the expected sensitivity
(and separately, expected bias) under that ROI's layer-preference
distribution:
\[
E[S]_r = \sum_{l=1}^{L} V_{r,l} \cdot S_l,
\]
where $V_{r,l}$ is the proportion of voxels in ROI $r$ best aligned with
layer $l$, and $S_l$ is layer $l$'s word-sensitivity (or, separately, its
recency bias). We restrict this analysis to the early third of layers for
sensitivity and the late third for bias, matching where each property is
concentrated (Figure~\ref{fig:expected_sens_bias}).

We do not find a consistent relationship across models. Expected
sensitivity in the early third declines from auditory to higher-order
regions for GPT-2 (17.9\% relative range across ROIs, monotonic), but is
flat for Llama-2 (15.8\%, no consistent direction) and reversed for Phi-2
(41.9\%, increasing toward higher-order regions). Expected recency bias in
the late third is comparably narrow in range across ROIs for all three
models (GPT-2: 11.5\%; Phi-2: 6.9\%; Llama-2: 6.3\%), consistent with our
interpretation of the late-layer signal as a domain-general confound
rather than a region-specific effect.

\FloatBarrier

One likely contributor to this inconsistency is that layer-wise
sensitivity itself is not monotonically decreasing with depth for all
models (Figure~\ref{fig:pos_averaged}). GPT-2's sensitivity
declines close to monotonically after an early peak. Phi-2 and Llama-2
both show a secondary rise in sensitivity at middle-to-late layers---most
pronounced for Llama-2, where sensitivity in layers 18--26 approaches
early-layer levels---meaning a simple ``early is sensitive, late is not''
story does not fully hold for these architectures at the layer level,
independent of any brain data. We therefore treat the content-sensitivity
account of early-layer/auditory-cortex alignment (Section\ref{explanations_predict_hierarchy})
as supported for GPT-2, and as an open question for Phi-2 and Llama-2,
rather than an architecture-general mechanism.

\begin{figure*}[t]
    \centering
    \includegraphics[width=1\linewidth]{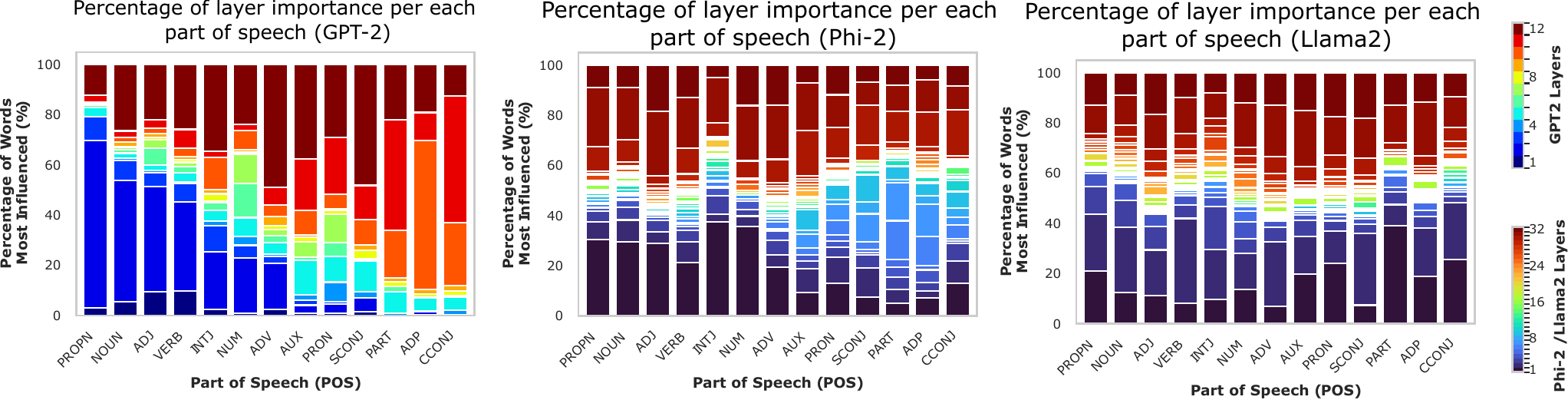}
    \caption{\textbf{Distribution of Layer importance per POS}  \label{fig:pos_dominance}}
\end{figure*}

\paragraph{Dominance-by-POS Control}\label{app:layer_dominance}
To better understand why early and late layers dominate the layer-importance profile, we examine layer dominance across different word categories. Figure~\ref{fig:pos_dominance} reports the distribution of dominant layers separately for each part-of-speech category. Across all categories, early and late layers remain the most frequently dominant, indicating that the bimodal layer-importance profile is not driven by a small subset of words or particular grammatical classes.

\end{document}